\documentclass[lettersize,journal]{IEEEtran}
\usepackage[english]{babel}
\usepackage{xcolor}
\usepackage{pifont}% http://ctan.org/pkg/pifont
\usepackage{booktabs} % For formal tables
\usepackage{multirow}
\usepackage{xspace}
\usepackage{enumitem}
\usepackage{balance}
\usepackage[multiple]{footmisc}
\usepackage{graphicx}
\usepackage{subcaption}
\usepackage{float}
\usepackage{hyperref}
\usepackage{booktabs}
\usepackage{longtable} % split long tables in multiple pages
\usepackage{supertabular}
\usepackage{caption}
\usepackage[autostyle]{csquotes}
\usepackage{listings}
\usepackage{tikz}
\usepackage{amsmath,amssymb,amsfonts}
\usepackage{algorithm}
\usepackage{algorithmic}

\usepackage{soul}
\usepackage{amssymb}% http://ctan.org/pkg/amssymb
\usepackage{pifont}% http://ctan.org/pkg/pifont
\usepackage{pgfplots} 
\usepackage[edges]{forest}
\usepackage{adjustbox} 
\usepackage{pifont}% http://ctan.org/pkg/pifont
\usepackage{textcomp}
\usepackage{adjustbox}
\usepackage{lscape}
\usepackage{tabularx}
\usepackage{color}
\usepackage[newcommands]{ragged2e}
\usepackage{colortbl}
\usepackage{array}
\usepackage{url}
\usepackage{rotating}
\usepackage{tabularray}
\usepackage{makecell}

\newcolumntype{C}[1]{>{\centering\arraybackslash\hspace{0pt}}p{#1}}
\def\BibTeX{{\rm B\kern-.05em{\sc i\kern-.025em b}\kern-.08em
   T\kern-.1667em\lower.7ex\hbox{E}\kern-.125emX}}

\title{FLAegis: A Two-Layer Defense Framework for Federated Learning Against Poisoning Attacks}

\author{Enrique M\'{a}rmol Campos, Aurora Gonz\'{a}lez-Vidal, Jos\'{e} L. Hern\'{a}ndez-Ramos and Antonio Skarmeta

\thanks{Enrique M\'{a}rmol, Aurora Gonz\'{a}lez-Vidal, Jos\'{e} L. Hern\'{a}ndez-Ramos and Antonio Skarmeta are with the University of Murcia, Spain. E-mail: \{enrique.marmol, aurora.gonzalez2, jluis.hernandez, skarmeta\}@um.es}}

\newcolumntype{M}[1]{>{\centering\arraybackslash}m{#1}}
\begin{document}

\markboth{IEEE Transactions on Dependable and Secure Computing}{FLAegis: A Two-Layer Defense Framework for Federated Learning Against Poisoning Attacks}
\maketitle

\begin{abstract}
Federated Learning (FL) has become a powerful technique for training Machine Learning (ML) models in a decentralized manner, preserving the privacy of the training datasets involved. However, the decentralized nature of FL limits the visibility of the training process, relying heavily on the honesty of participating clients. This assumption opens the door to malicious third parties, known as Byzantine clients, which can poison the training process by submitting false model updates. Such malicious clients may engage in poisoning attacks, manipulating either the dataset or the model parameters to induce misclassification. In response, this study introduces FLAegis, a two-stage defensive framework designed to identify Byzantine clients and improve the robustness of FL systems. Our approach leverages symbolic time series transformation (SAX) to amplify the differences between benign and malicious models, and spectral clustering, which enables accurate detection of adversarial behavior. Furthermore, we incorporate a robust FFT-based aggregation function as a final layer to mitigate the impact of those Byzantine clients that manage to evade prior defenses. We rigorously evaluate our method against five poisoning attacks, ranging from simple label flipping to adaptive optimization-based strategies. Notably, our approach outperforms state-of-the-art defenses in both detection precision and final model accuracy, maintaining consistently high performance even under strong adversarial conditions.

\end{abstract}

\IEEEpeerreviewmaketitle

\section{Introduction} \label{sec:introduction}
Federated Learning (FL) has emerged as a powerful technique that has captured the attention of numerous researchers \cite{mcmahan2017communication, yang2019federated}. This approach was devised to address contemporary challenges concerning privacy and the adherence to the General Data Protection Regulation (GDPR) principles \cite{goddard2017eu}, enabling model training across decentralized data sources that cannot be centrally collected due to privacy, legal, or practical constraints \cite{yang2019federated}. FL involves multiple clients training a common Machine Learning (ML) model, each one training the model independently using its own data. These clients periodically transmit locally computed model updates (i.e., weights or gradients) to a central server. Specifically, the server aggregates incoming updates using an aggregation function, and sends the resulting global model back to the clients for further local training. This decentralized approach allows clients to benefit from collaborative training without sharing their raw data, mitigating the need for centralized data storage.

However, as the server remains agnostic to the clients' local training processes, this setup creates an opening for malicious actors aiming to disrupt the FL pipeline. Specifically, so-called Byzantine clients \cite{blanchard2017machine} can compromise the training process through poisoning attacks \cite{barreno2006can}. These attacks can take the form of \textit{data poisoning}, in which the training data is intentionally manipulated, or \textit{model poisoning}, where the model parameters are maliciously altered before the aggregation. The objective is to craft poisoned updates that, when aggregated, cause global model degradation or targeted misclassification in benign clients. Previous work \cite{bagdasaryan2020backdoor} has demonstrated that FL systems are highly vulnerable to such attacks, even when the number of malicious participants is small. Consequently, the server, being blind to the internal state or data of clients, becomes a weak point for adversarial influence.

In response to this challenge, the research community has explored a range of defenses against poisoning attacks in FL \cite{tian2022comprehensive}. Two primary lines of defense have emerged \cite{xia2023poisoning}. The first leverages robust aggregation functions, a special type of aggregation function that aims to mitigate the impact of malicious updates without explicitly identifying adversaries. Although simple and scalable, such methods typically degrade under high Byzantine ratios  \cite{fang2020local}. The second approach employs unsupervised clustering or anomaly detection to identify and exclude malicious clients from the aggregation process. While more precise, these methods can be sensitive to noise and require careful tuning \cite{sattler2020byzantine}. 

To address these limitations, we introduce \textit{FLAegis}, a two-layer defense framework that first detects and excludes potentially malicious clients before aggregation, and then applies a robust aggregation function to reduce the residual impact of undetected adversaries. Our methodology involves processing client-side model updates using Symbolic Aggregate approXimation (SAX) \cite{lin2003symbolic}. SAX discretizes numerical sequences into symbolic representations, effectively amplifying structural differences between benign and adversarial behaviors. Then, for evaluating model similarity among clients, we leverage the cosine similarity function \cite{KOTU2019343}. Based on these pairwise similarities, we construct a similarity matrix and apply spectral clustering \cite{simovici2021clustering}, which allows grouping of clients without predefining the number of clusters. This represents a key advantage in adversarial contexts where the number of attackers is unknown. When the algorithm detects more than one cluster, that indicates the presence of malicious clients. We make a final clustering of 2 groups, and we flag the smallest as potentially malicious. Despite these measures, there may still exist instances where certain byzantine clients circumvent all filters, potentially infiltrating the aggregation process and consequently affecting the global model. To mitigate this, we incorporate a second defense layer using the Fast Fourier Transform (FFT), inspired by our prior work \cite{fourier}, to perform robust aggregation in the frequency domain. This function filters out frequency components associated with anomalous patterns, reducing the effect of outlier updates.

Unlike our previous work \cite{fourier}, which relied solely on robust aggregation without identifying attackers, FLAegis combines detection and mitigation in a unified architecture and extends the evaluation to a broader range of poisoning attacks and defense mechanisms. In this direction, through the reduction of the number of malicious clients with the detection layer, we overcome the disadvantages of robust aggregation functions against high Byzantine ratios. The resulting approach significantly improves performance under high adversarial presence. Our contributions can be summarized as follows:

\begin{itemize}
    \item A two-stage defense pipeline that dynamically adapts to both benign and adversarial scenarios without degrading model performance. 
    \item Integration of SAX preprocessing to amplify dissimilarity between benign and adversarial updates via symbolic abstraction of weight sequences.
    \item  Additional defense layer using FFT-based aggregation, which preserves model performance even when some attackers evade detection.
    \item Empirical validation showing that our framework outperforms state-of-the-art defenses in both model accuracy and malicious client detection.
\end{itemize}

The remainder of this paper is organized as follows: Section \ref{sec:Preeliminaries} details the fundamental concepts supporting our work. Section \ref{sec:Related} presents an overview of existing research in this domain. Section \ref{sec:Methodology} delineates the procedural aspects of our methodology. In Section \ref{sec:Results}, we present the outcomes and findings of our method, encompassing the framework, datasets, and parameters involved. Finally, Section \ref{sec:Conclusions} offers concluding remarks.

\section{Preliminaries}\label{sec:Preeliminaries}
This section introduces the core concepts and techniques of the proposed FLAegis framework. We provide a concise overview of poisoning attacks in FL, followed by the fundamental principles of SAX, spectral clustering, and the Fast Fourier Transform (FFT). These components form the methodological basis for the proposed system, enabling both detection and mitigation of adversarial behaviors within the FL process.

\subsection{Poisoning attacks in FL}
FL enables collaborative model training across multiple decentralized clients while preserving data privacy, as originally introduced in \cite{mcmahan2017communication}. A typical FL setting comprises a set of \textit{clients} and an \textit{aggregator} (or \textit{server}). Each client holds a unique dataset that remains inaccessible to other clients throughout the FL process. The clients train a common ML model and periodically send local model updates (i.e., weights) to the server. The aggregator combines these updates using an \textit{aggregation function} and returns the global model to the clients for further training. The most common aggregation function called FedAvg computes the average of client weights \cite{mcmahan2017communication}. Since the server lacks visibility into clients' local training processes or data, this setup introduces a vulnerability: malicious clients can manipulate their updates to poison the global model without being easily detected.

Poisoning attacks aim to degrade model performance by injecting malicious data or tampering with weight updates \cite{barreno2006can}, often leading to misclassification. In the context of FL, such attacks not only compromise the adversary's local model but can also propagate their effects to benign clients through the aggregation step. There are two types of poisoning attacks: data poisoning, and model poisoning. Data poisoning involves polluting the local data of a client, whereas model poisoning entails changing the weights produced after training, so that clients send corrupted weights to the server. Furthermore, data poisoning attacks fall into two categories: clean-label and dirty-label  \cite{lyu2020threats}. Clean-label attacks modify only the inputs while keeping the labels unchanged, making them more difficult to detect. In contrast, dirty-label attacks modify both inputs and labels. In the case of model poisoning attacks, they can be further categorized as untargeted or targeted \cite{tian2022comprehensive}. Untargeted attacks aim to cause the global model to misclassify broadly with minimal perturbation, without targeting any specific class. Targeted attacks aim to induce misclassification for specific classes or inputs. Indeed, these attacks could be formulated as a multi-task problem since they force the global model to perform abnormally on specified samples while ensuring its legitimate function on other benign samples. These attack categories define the threat model (see Section \ref{sec:threat_model}) under which our FLAegis framework is evaluated.

\subsection{Symbolic Aggregate approXimation}

A time series is a sequential dataset representing recorded values of a phenomenon over time. Time series analysis captures both variability and structural patterns in sequential data, with applications in domains ranging from finance to medicine and cyber-physical systems \cite{rhif2019wavelet}. Given that the sequential nature of weight updates during the training process reflects the temporal dynamics of the model, we treat the sequence of model weights as a time series, enabling temporal analysis techniques to be applied in the FL setting. Typical time series data tasks that involve series classification, clustering, rule extraction, and pattern querying can be enhanced through data aggregation and pattern extraction methods \cite{chaovalit2011discrete}. These preprocessing steps help reduce computational cost while revealing hidden structure \cite{gonzalez2018beats}, in both univariate and multivariate scenarios \cite{gonzalez2024multibeats}. Time series representations can be data-adaptive and non-data-adaptive \cite{lin2003symbolic}. Non-adaptive techniques apply fixed transformations regardless of the dataset, whereas adaptive ones tailor the representation to the statistical properties of the data.

SAX \cite{lin2003symbolic} is a symbolic data-adaptive method that transforms real-valued time series into symbolic representations (strings). We employ SAX  to transform model weights into symbolic form. SAX is based on Piecewise Aggregate Approximation (PAA) \cite{keogh2001dimensionality}, which divides a series into equal-length segments and computes the mean of each. These means are then mapped to symbols based on predefined breakpoints, allowing the original time series to be encoded compactly. An important property of SAX is that distance measures in the symbolic space lower-bound their counterparts in the original space, enabling efficient similarity computations.

In our approach, we apply SAX to each client's weight update by first flattening the weight matrices into one-dimensional vectors, allowing them to be treated as time series. Each value is mapped to a symbol based on the segment it falls into. For instance, a transformed sequence might result in a symbolic representation such as ``$CBCCBBCCACBBBAAAABAA$'', where each character corresponds to a quantized region of the original value range. This process discretizes the weights, filtering out minor fluctuations while amplifying significant deviations, such as those introduced by adversarial manipulation. As a result, SAX enables robust and compact comparison between client updates, which is crucial for detecting anomalous behavior in the presence of poisoning attacks.

\subsection{Spectral clustering}

Given a set of points $P$ and a similarity matrix $W$ measuring similarities among these points, spectral clustering aims to categorize the dataset into distinct subsets characterized by high intra-cluster similarity and low inter-cluster similarity. Traditional methods like K-means \cite{lloyd1982least} and Gaussian mixture models \cite{yu2015gaussian} perform well when clusters are convex or linearly separable. However, they often fail in scenarios with non-convex or complex geometries, such as interleaved spirals, where spectral clustering is more effective due to its nonparametric nature and lack of assumptions about cluster shape. Notably, the main property that motivates our use of spectral clustering, (as described in Section \ref{sec:Methodology}), is that, unlike K-means and Gaussian mixture models, it does not require the number of clusters to be predefined \cite{liu2018spectral}; the number of clusters can be inferred from the data structure itself. This property is crucial in adversarial settings such as ours, where the number of malicious clients is unknown and might vary across rounds.

Spectral clustering operates by representing the dataset as a weighted graph, where each point is a node and edge weights reflect pairwise similarities (stored in the matrix $W$). Several methods exist for computing this matrix $W$ \cite{von2007tutorial, bach2003learning}. Subsequently, the graph Laplacian matrix \cite{simovici2021clustering} $L$ and the degree matrix $D$ are calculated, with $D$ being a diagonal matrix where each diagonal entry represents the sum of similarities for a given node. In this case, the Laplacian matrix will be $L=D-W$, and the subsequent step involves computing eigenvectors $(e_1,\dots, e_k)$, with $k$ denoting the final number of clusters. These eigenvectors $(e_1,\dots, e_k)$ form a matrix $F$, whose rows are treated as new representations of the original points. Finally, a simple clustering algorithm, such as K-means, is applied to the rows of $F$ to produce the final partitioning.

 In our framework, spectral clustering is applied to a similarity matrix derived from clients’ weight updates after SAX transformation. This allows to detect malicious groups based on their statistical divergence from benign clients, even without knowing how many attackers are present.

\subsection{Fast Fourier Transform}
The use of the Fast Fourier Transform (FFT) for robust aggregation in FL was introduced in our previous work \cite{fourier}. The Fourier transform (FT) enables us to extend the theory of the Fourier series \cite{tolstov2012fourier} into functions defined in all the real numbers. The FT projects a function into its frequency domain, allowing for the calculation of the density function of the processed data. In practice, we employ the FFT, an efficient algorithm that reduces the computational complexity from $O(N^2)$ to $O(N \log N)$ for a signal of length $N$. In the context of aggregation robustness, the main reason for choosing the FFT is its ability to identify and amplify consistent patterns across clients, even in the presence of adversarial noise. Specifically, in a coordinate-wise manner, at a layer $l$, we take the element $i$ of each client to form the vector $V_{i,l}$. For this vector, we use the FFT to calculate the density function from its frequency components, and then select the point with the highest value on the density function, which corresponds to the point of the highest frequency. Since the FFT is easily invertible, we can retrieve the associated value in the original domain without additional overhead. Repeating this process for the rest of the elements and layers provides the final aggregated model parameters distributed by the server. Experimental results from \cite{fourier} show that this frequency-based strategy outperforms several state-of-the-art aggregation methods under different adversarial conditions.
\section{Related work}\label{sec:Related}

This section is structured in two parts. First, we provide a comprehensive review of existing approaches to mitigate poisoning attacks, with a particular focus on methods that aim to identify and exclude malicious clients. In the second part, we analyze a selection of techniques that demonstrate robust and competitive performance in realistic adversarial settings. These selected methods are included as evaluation baselines, serving as comparative references in the experimental analysis presented in Section~\ref{sec:Results}.

\subsection{Existing defense strategies}

In recent years, the resilience of FL systems against poisoning attacks has received significant interest. Various approaches have emerged to address this issue. One approach involves employing robust aggregation functions capable of withstanding malicious updates. A diverse range of functions has been explored, such as median \cite{yin2018byzantine}, Krum \cite{blanchard2017machine}, trimmed mean \cite{yin2018byzantine}, FoolsGold \cite{fung2018mitigating}, FLtrust \cite{cao2020fltrust}, and FFT \cite{fourier}. These functions provide a layer of protection but often incur trade-offs in terms of model performance. Furthermore, these functions present two main problems. Firstly, certain functions require the specification of the number of malicious clients (e.g., Krum, trimmed mean), which is hard to know in real-case scenarios. Secondly, their efficacy diminishes significantly when confronted with a high percentage of malicious clients, resulting in a substantial decay in performance \cite{cao2020fltrust}.

Consequently, several studies aim to identify and isolate malicious clients before aggregating the updates from the remaining benign participants using standard functions. Notably, these works employ diverse techniques to detect malicious clients and protect the global model. For instance, \cite{li2020learning} uses a Variational Autoencoder to classify clients as benign or malicious. However, this approach assumes access to a small, trusted validation dataset at the server, which may limit applicability in highly non-IID scenarios, and adds significant training overhead due to the pre-training stage required prior to FL execution. A common technique involves clustering methods used to differentiate between benign and malicious clients. For example, \cite{zhang2022fldetector} introduces FLDetector, which leverages the Hessian matrix of weights to predict subsequent client weights. By measuring the disparity between the real and the predicted weights of each client, FLDetector utilizes gap statistics to determine the number of client clusters. Subsequently, if the count exceeds one, they employ k-means to partition clients into two clusters: benign and malicious ones. However, the cumulative computational cost of these steps can render deployment on resource-constrained devices impractical.  

Another prevalent approach relies on model similarity metrics such as cosine similarity. In \cite{feng2023sentinel}, the authors introduce Sentinel, a defense strategy against poisoning attacks in FL. Using cosine similarity, Sentinel filters out clients exhibiting lesser similarity compared to others. Subsequently, the authors calculate the loss of remaining clients, eliminating those with higher loss. Finally, they protect the system from those clients that evade the previous filters by normalizing their gradients. However, although this method includes a step to mitigate the impact of large-norm gradients, it may still struggle when gradients have similar norms but opposite directions. Moreover, the effectiveness of Sentinel against advanced attacks is not thoroughly validated. Furthermore, \cite{li2021lomar} presents LoMar, an algorithm designed to defend the FL process against poisoning attacks. Their algorithm is divided into two parts, first, authors divide the model weights using the $k$ nearest neighbors and then apply kernel density estimation (KDE) to calculate what they call the client's malicious factor. In the second part, they set a threshold to assess which clients are malicious. Finally, FedDMC \cite{mu2024feddmc} reduces the dimension of the clients' weights to enhance the dissimilarity between malicious and benign clients. Next, they cluster the clients using a binary tree classification to separate benign from malicious clients. Finally, FedDMC calculates a trust score based on the times the clients were classified as benign or malicious to make the final decision. 

Moreover, studies like \cite{xu2022byzantine, wan2022shielding} suggest that naïve direct application of cosine similarity may yield similar results between benign and malicious clients, particularly when adversaries craft updates to mimic benign directions. Hence, \cite{wan2022shielding} modifies the input for cosine similarity in the method they proposed, MAB-RFL. This approach filters out potential attackers based on past attacks in previous rounds. Additionally, they differentiate between Sybil and non-Sybil attacks \cite{douceur2002sybil}. For Sybil attacks, they create a graph based on the cosine similarity to detect Sybil clients (i.e., when an attacker generates multiple fake data points or models to manipulate the training process or outcomes), and for non-Sybil attacks, they cluster the clients based on gradient similarity. However, the assumption that the same attackers are active in every round undermines the effectiveness of this initial filter. Additionally, the gradient normalization techniques used in \cite{feng2023sentinel} and \cite{wan2022shielding} may slow convergence and distort updates from benign clients, potentially harming final accuracy. Lastly, in \cite{xu2022byzantine}, the authors introduce SignGuard, a framework devised to detect malicious clients. They employ a two-filter approach: initially, by using gradient norms, they eliminate those exceeding the median value. Subsequently, they employ mean statistics to cluster clients into malicious and benign categories. Additionally, they extend their methodology by incorporating cosine similarity into their analysis, and evaluate it under both IID and non-IID scenarios.

In contrast, our approach addresses the main limitations identified in the previous works. First, we adopt methods that are less computationally demanding compared to normalization-based or high-dimensional clustering techniques. Second, to address the limitations of cosine similarity highlighted by \cite{xu2022byzantine, wan2022shielding}, we apply SAX as a preprocessing step before the cosine similarity, augmenting the dissimilarity between malicious and benign updates and thus improving the quality of the similarity matrix used in the clustering phase. Third, to provide a resilient defense against malicious clients that evade the detection stage, we integrate FFT-based robust aggregation within the FL pipeline. Our approach neutralizes malicious contributions without degrading benign updates, while avoiding the convergence delays related with normalization-based techniques (\cite{feng2023sentinel, wan2022shielding}). Moreover, we conduct extensive evaluations against a wide variety of attacks, including label flipping and less common ones such as LIE, STATOPT, mirror, min-max, and min-sum, in a non-IID data setting. These results show that our framework consistently outperforms existing defenses in both resilience and accuracy under challenging adversarial conditions.

\subsection{Selected methods for comparison}\label{sec:related_methods_for_comparison}
Building upon the previous analysis of the state of the art approaches, we identify a subset of representative methods whose designs are robust or widely adopted, making them suitable baselines for comparative evaluation. In this regard, to evaluate their effectiveness, we include them in our experimental comparison in Section~\ref{sec:Results}, where their performance is assessed alongside our proposed model FLAegis. These works are: SignGuard, FedDMC, and LoMar, which are further described below. 

\subsubsection{SignGuard}

SignGuard is introduced in \cite{xu2022byzantine} as a method for filtering clients and selecting only the benign ones. To achieve this, the method leverages client gradient information, employing two stages of filters. The first filter is based on the gradient norm. It computes the median of the gradient norms, and flags clients whose gradient norms fall outside user-defined lower and upper bounds in Eq. \ref{eq:singguard1} as potentially malicious. In this equation, $g_r^k$ denotes the gradient of client $k$ at round $r$, $M$ is the median of the norms, and $L$ and $R$ are constants set by the method’s configuration.

\begin{equation}\label{eq:singguard1}
    L\leq \frac{||g_r^k||}{M}\leq R
\end{equation}

Subsequently, the second filter is applied to identify clients whose gradient sign vectors are inconsistent with the majority. The method applies Mean-shift clustering to categorize clients based on their gradient sign vectors, and the smallest cluster is identified as the cluster of malicious nodes. Finally, the union of the two filtered sets is considered malicious, and aggregation is performed using Eq. \ref{eq:singguard2}. $S_t'$ is the subset of clients considered in that round, and $\min \left( 1, \frac{M}{\|g_r^{k}\|} \right)$ serves to limit the impact of gradients with large norms, preserving only their direction.

\begin{equation}\label{eq:singguard2}
    \tilde{g}_r = \frac{1}{|S_t'|} \sum_{i \in S_t'} g_r^{k} \cdot \min \left( 1, \frac{M}{\|g_r^{k}\|} \right)
\end{equation}

\subsubsection{FedDMC}
FedDMC is proposed in \cite{mu2024feddmc} as an FL framework to defend against poisoning attacks. The architecture consists of three main modules: i) a dimensionality reduction module to amplify dissimilarities between benign and malicious clients while also reducing computational complexity, ii) a binary tree-based classification module to identify malicious clients, and iii) a self-ensemble detection correction module that stabilizes detection by combining past and current classification results. Initially, the dimensionality reduction module implements Principal Component Analysis (PCA) to project model weights into a lower-dimensional space. This transformation reduces computational complexity and highlights discriminative features for the subsequent clustering stage. Next, a binary tree is constructed based on pairwise Euclidean distances among the projected weights. Each leaf corresponds to a client. Internal nodes recursively split the dataset into two subclusters maximizing inter-cluster distance. A child cluster is marked as malicious if it fails to meet a minimum leaf count threshold. Specifically, they construct a binary clustering tree \(T\) over the projected weights \(\{\tilde w_i\}_{i=1}^n\) by applying hierarchical agglomerative clustering using Euclidean distance.
\begin{equation}
T \;=\; \mathrm{HAC}\bigl(\{\tilde w_i\}_{i=1}^n\bigr).
\end{equation}

Then, the current node is initialized to the root of \(T\) and sets the detected‑outlier set \(\mathcal{O}\) to be empty.
\begin{equation}
r_0 = \mathrm{root}(T),
\quad
\mathcal{O}_0 = \varnothing.
\end{equation}

At each iteration \(t=0,1,2,\dots\), the current node \(r_t\) is split into its two child clusters \((L_t,R_t)\).
\begin{equation}
(L_t,\,R_t) \;=\;\mathrm{Split}(r_t).
\end{equation}

Next, the set of detected outliers is updated by adding whichever child is declared an outlier (if any).
\begin{equation}
\mathcal{O}_{t+1} \;=\;
\begin{cases}
\mathcal{O}_t\cup\{L_t\}, & \text{if }L_t\text{ is an outlier},\\[4pt]
\mathcal{O}_t\cup\{R_t\}, & \text{if }R_t\text{ is an outlier},\\[4pt]
\mathcal{O}_t,            & \text{otherwise}.
\end{cases}
\end{equation}

Then, the next node \(r_{t+1}\) is selected as the child that was not marked as an outlier (or remains if neither was).
\begin{equation}
r_{t+1} =
\begin{cases}
R_t, & \text{if }L_t\text{ was marked outlier},\\[4pt]
L_t, & \text{if }R_t\text{ was marked outlier},\\[4pt]
r_t, & \text{otherwise}.
\end{cases}
\end{equation}

Iteration stops as soon as at least half of the clients have been labeled as outliers.

Finally, the binary detection labels \(S=\{s_i\}_{i=1}^n\) are created, where each
\(s_i=1\) if client \(i\) was ever assigned to \(\mathcal{O}\), and \(0\) otherwise.
\begin{equation}
s_i =
\begin{cases}
1, & i\in \mathcal{O}_{t^*},\\[4pt]
0, & \text{otherwise},
\end{cases}
\quad
S = \{s_i\}_{i=1}^n,
\end{equation}
where \(t^*\) is the first iteration satisfying the stopping criterion.

Finally, after this classification, the server has a historical vector $S_r$ of the times each client was classified as benign or malicious at round $r$. For each round $r$, the server computes a weighted average by the parameter $\alpha$ between the current result and the past results to decide whether this client is considered malicious or benign according to the next equation:

\begin{equation}
S_r=
\begin{cases}
0.5, & \text{if } r =0\\[4pt]
\alpha S_{r-1}+ (1-\alpha)S_r, & \text{if }r>0.
\end{cases}
\end{equation}

\subsubsection{LoMar}

LoMar is proposed in \cite{li2021lomar} as a two-phase defense algorithm designed to address poisoning attacks against FL. The first phase focuses on calculating malicious factors $F(i)$, which quantify the degree of maliciousness of the clients. To calculate them, LoMar first computes the k-nearest neighbors for each client and estimates the local distribution of the client's output on each label dimension. This is done using the following formula:
\begin{equation}
    \tilde{q}(u_i^r) = \frac{1}{k} \sum_{j=1}^k K\!\left(\frac{u_i^r - u_{i,j}^r}{h^r}\right),
\end{equation}

where \(K(\cdot)\) is the kernel function (Gaussian function in this case) and \(u_{i,j}^r\) denotes the \(r\)-th output from the \(j\)-th neighbor in \(\mathcal{U}_i\), and $h^r$ is the bandwidth of $K(\cdot)$.  The maliciousness factor on label r is then defined as:

\begin{equation}
    F(i)^r = \frac{\sum_{j=1}^k \tilde{q}\bigl(u_{i,j}^r\bigr)}{k\,\tilde{q}\bigl(u_i^r\bigr)},
\end{equation}
and consequently, $F(i) = \prod_r F^r(i)$

In the second phase, clients are classified based on whether their score $F(i)$ falls below a threshold $\epsilon$. If $F(i)$ $<$ $\epsilon$, then it is malicious, otherwise, not. To estimate $\epsilon$, the following inequality is used to bound the probability that a benign client's score exceeds a threshold $\varepsilon_m$:

\begin{equation}\label{eq:lomar2}
    P\bigl(F(i) > \varepsilon_m\bigr) \le \exp\left(-\frac{4\pi(\varepsilon_m - 1)^2 (k + 1)^2 \hbar}{k(2k + \varepsilon_m + 1)^2 V^2}\right)
\end{equation}

The final threshold is set as $\epsilon= min\{1, \epsilon_m \}$  where 1 serves as an upper bound by definition. Only the clients not flagged as malicious are included in the final aggregation step.

Based on the description of previous approaches, we now present our defense strategy, which combines efficient client filtering and robust aggregation to address poisoning attacks in FL. The next section details the assumptions, system components, and each step of our proposed FLAegis approach.

\section{FLAegis Threat Model and Framework}\label{sec:Methodology}
This section introduces the FL setting considered in this work, along with the assumptions that define our threat model. We then describe the complete architecture and operational steps of FLAegis to identify and mitigate poisoning attacks.

\subsection{Adversary model}\label{sec:threat_model}
\begin{figure*} [!ht]
	\centering
		\includegraphics[width=1.6\columnwidth]{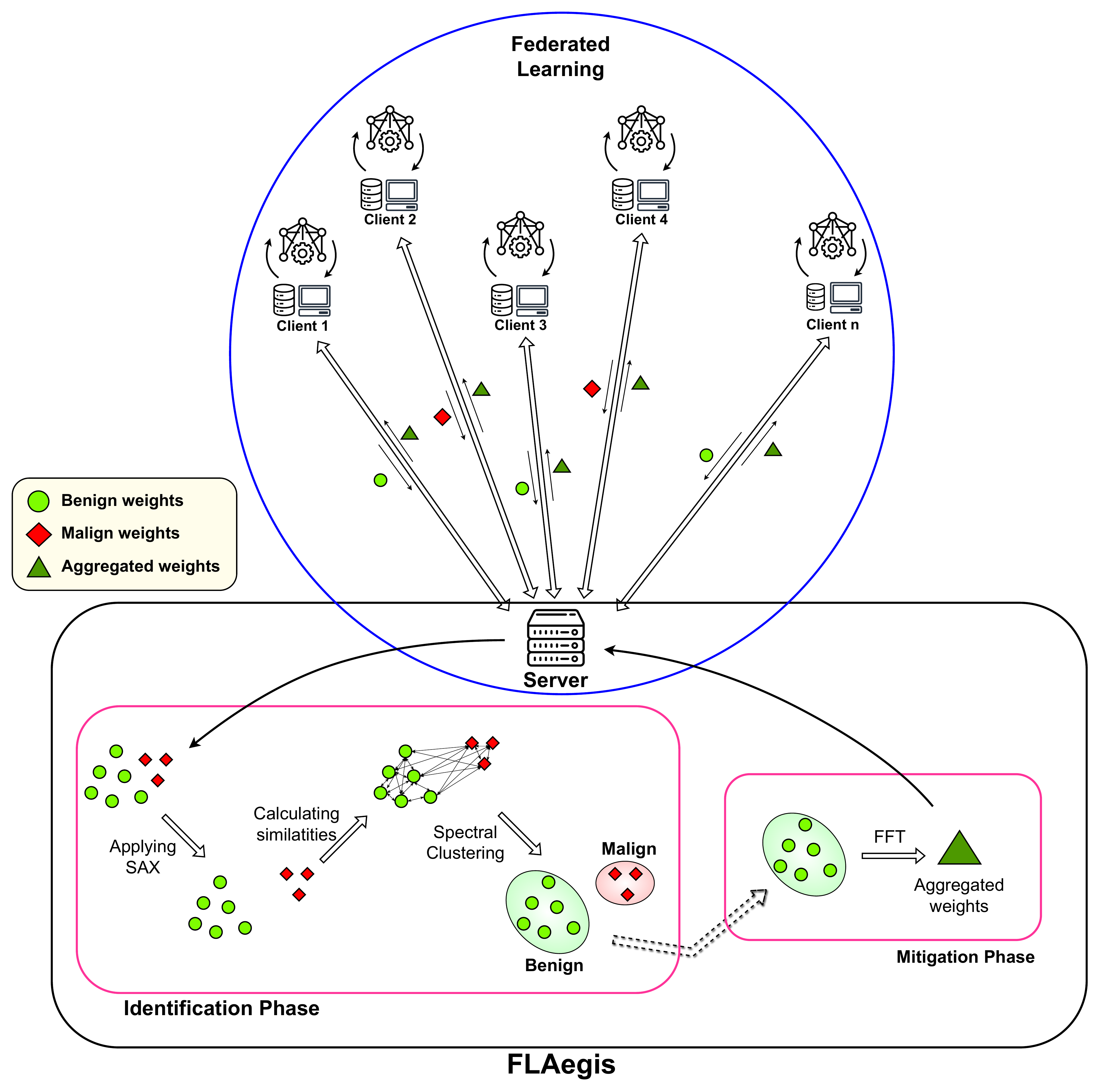}
	\caption{Overview of the FL setting. Benign clients send clean model updates (circles), while malicious clients may collaborate or act independently to send poisoned updates (diamonds). The server filters and removes detected malicious updates before applying FFT for robust aggregation (triangles), then sends the global model to the clients.}
	\label{fig:FLsetup}
\end{figure*}

We consider the FL setup introduced in Section \ref{sec:Preeliminaries} to detail the threat model, as illustrated in Fig.~\ref{fig:FLsetup}. Our setup builds upon our previous work \cite{fourier}, but introduces a key enhancement in which the server can \textit{identify and exclude malicious clients prior to aggregation}. Indeed, while \cite{fourier} applies FFT as the aggregation approach to attenuate the influence of adversarial updates \textit{without explicit detection}, our method integrates a preliminary identification phase to enhance robustness.

\begin{figure*}
    \centering
    \begin{subfigure}{0.45\textwidth}
        \centering
        \includegraphics[width=\columnwidth]{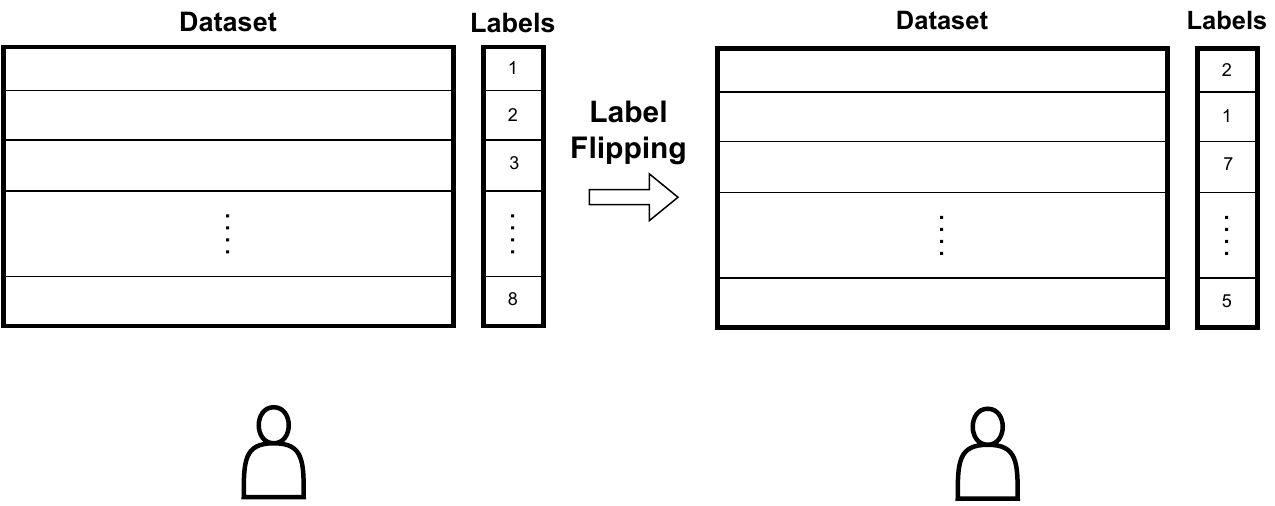}
        \caption{Label flipping}
        \label{fig:labelf}
    \end{subfigure}
    \hfill
    \begin{subfigure}{0.45\textwidth}
        \centering
        \includegraphics[width=0.7\columnwidth]{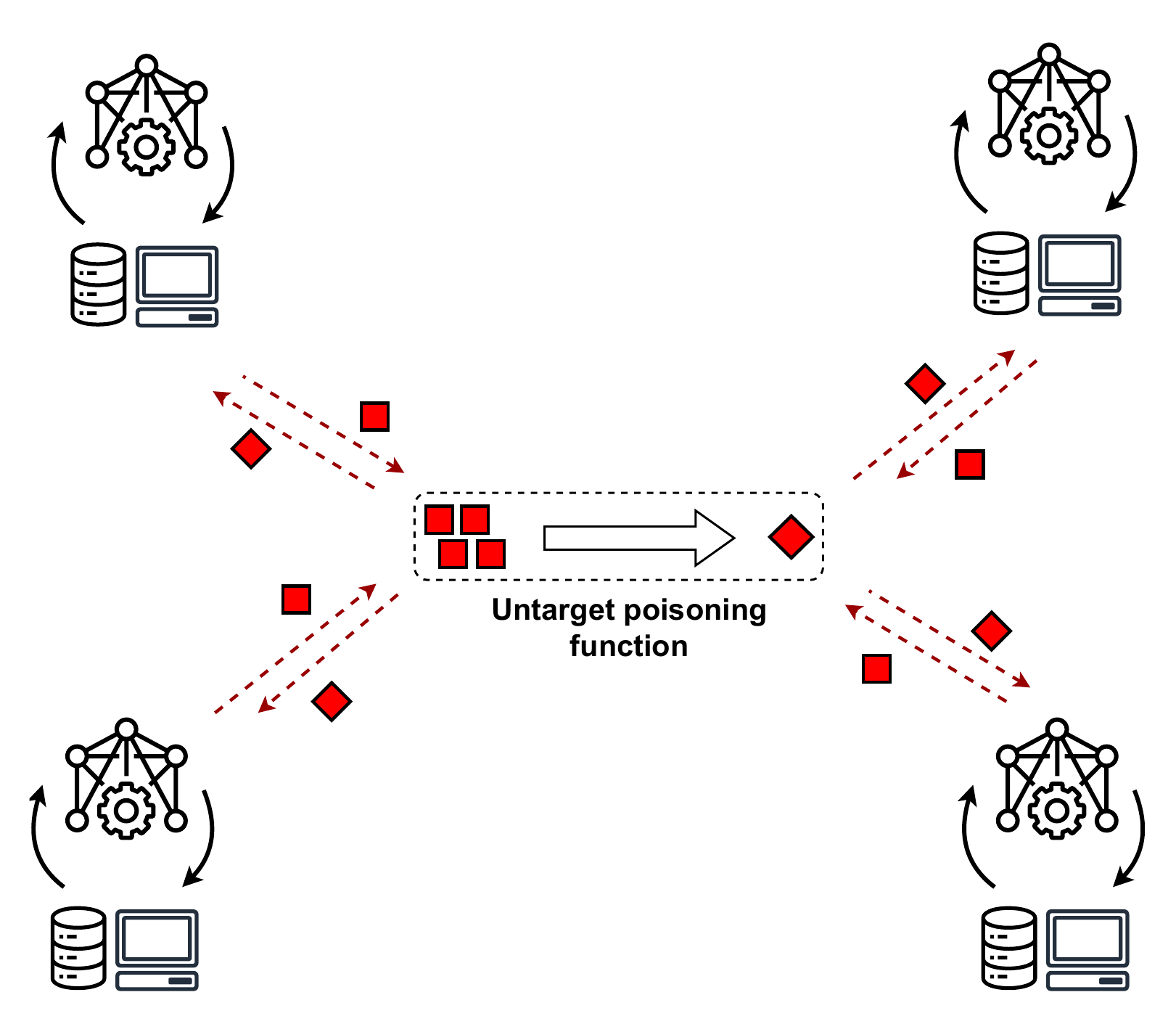}
        \caption{Untargeted attack}
        \label{fig:figura2}
    \end{subfigure}
    \caption{Scheme of the two types of poisoning attacks. (a) Label flipping: dataset labels are randomly altered. (b) Model poisoning: malicious clients collaborate to craft harmful model updates and disrupt training.}
    \label{fig:untar}
\end{figure*}

The considered FL system comprises $K$ clients collaboratively training a model via a central aggregator. Although we adopt a single-aggregator setup, the proposed method generalizes to decentralized or hierarchical FL architectures \cite{lalitha2018fully}. We assume that a subset of clients (denoted by $M$) are Byzantine adversaries attempting to corrupt the learning process. As in prior work \cite{xu2022byzantine, wan2022shielding}, we assume $M < K/2$, which ensures the feasibility of clustering-based detection. During training, as shown in Fig.~\ref{fig:FLsetup}, these adversaries generate and send poisoned model updates (red diamonds) instead of legitimate ones (green circles), aiming to distort the aggregated model and hinder convergence. Furthermore, the malicious updates are generated through either data poisoning or model poisoning techniques, as depicted in Fig.~\ref{fig:untar}. In data poisoning (e.g., label flipping), clients manipulate training labels to degrade model generalization. In model poisoning, malicious clients coordinate to craft adversarial updates that maximize disruption to the global model. To cope with such behavior, the server identifies and discards malicious clients before applying the FFT to the remaining benign updates, as shown in Fig.~\ref{fig:FLsetup}. We assume the aggregator is trustworthy and not compromised. During evaluation, each malicious client $m \in M$ performs one of five attacks analyzed in this study, detailed in the next section.

Our threat model builds upon the systematization of threat models for poisoning attacks in FL proposed by \cite{shejwalkar2022back}. This framework describes different aspects related to the characteristics of the attackers, including their objectives, knowledge, and capabilities:

\begin{itemize}
    \item \textbf{Objective of the adversary}: the goal of Byzantine clients is to degrade the global model's performance by injecting poisoned updates or corrupting local datasets. Consistent with \cite{baruch2019little}, we focus on \textit{indiscriminate} attacks, which aim to broadly reduce accuracy rather than target specific classes or instances.
    
    \item \textbf{Knowledge of the adversary}: adversaries in local model poisoning scenarios can access their own and potentially other malicious clients' updates, but not those from benign clients. In data poisoning, attackers manipulate only their local dataset. Therefore, following the categorization described by \cite{shejwalkar2022back}, we assume \textit{whitebox} knowledge regarding the models produced by the clients in each training round. Moreover, although it could be considered in future work, we assume that the attackers do not have information related to the aggregation function used.
    
    \item \textbf{Capabilities of the adversary}: malicious clients can manipulate local data (in the case of data poisoning) or alter model updates during training (in model poisoning). This can occur through direct device access or interception mechanisms like man-in-the-middle attacks. Furthermore, these attacks follow an \textit{online} mode, i.e., these attacks are launched in every training round they are executed continuously throughout the training process. In addition, the attackers have the ability to collude with each other by exchanging their local updates to achieve a greater impact.
\end{itemize}

\subsection{FLAegis Detection and Mitigation Pipeline}
This section details the workflow of our malicious client detection system, FLAegis. As illustrated in Algorithm \ref{alg:methodology} and Fig. \ref{fig:FLsetup}, the system is composed of two main phases: an identification phase, which detects and isolates malicious clients, and a mitigation phase, which minimizes the residual impact of undetected adversaries on the global model.

\subsubsection{Identification Phase}
We start from the previously defined setting, where there are $M$ malicious clients performing poisoning attacks against the global model. Our goal is to provide a defense system that can isolate these malicious updates while preserving the performance of the global model. As shown in Algorithm \ref{alg:methodology}, first, we apply SAX to each client's weight vector. In our implementation, the SAX transformation discretizes the input sequences by dividing the value range into 45 equidistant bands, assigning one unique symbol per band. This fine-grained symbolic representation allows for capturing subtle but consistent variations introduced by adversarial behaviors. Subsequently, we construct the similarity matrix $M$ based on the cosine similarity \cite{KOTU2019343} of the transformed weights. SAX is applied to better differentiate weights between clients, addressing prior findings \cite{xu2022byzantine, wan2022shielding} that cosine similarity alone may be insufficient for measuring model similarity. By design, SAX amplifies small differences in the weight vectors, making cosine similarity more discriminative. This allows malicious weights to be mapped into separate symbolic bands, distinguishing them more clearly from benign ones. Once the similarity matrix has been calculated, we apply spectral clustering. If more than one cluster is identified, it suggests the presence of anomalous clients whose weights differ significantly from the rest. As spectral clustering utilizes K-means for final cluster labelling, we use K-means to divide the clients into two clusters: benign and malicious. Finally, we select the smaller cluster as containing the malicious clients, as previously described. It is worth noting that this decision rule (selecting the smaller cluster as the malicious one) is based on the common assumption that $M < K/2$, which is necessary to avoid misclassifying the benign majority. This heuristic aligns with prior works and is reflected in lines 13–17 of Algorithm \ref{alg:methodology}.

\subsubsection{Mitigation Phase}
Despite the identification phase, as noted in \cite{feng2023sentinel, xu2022byzantine}, it is possible that some malicious clients evade detection in certain rounds.  This residual threat can still degrade the model’s performance unless further action is taken. To address this, we incorporate a second defense layer by applying the robust aggregation strategy described in \cite{fourier}, which serves as a safeguard against undetected malicious clients. Unlike approaches such as \cite{feng2023sentinel}, which normalize weights and may affect model convergence, we directly apply the robust aggregation to the filtered weights without modifying their scale. While robust aggregation methods may exhibit limitations when used in isolation, their integration as a secondary defense (after most malicious clients have been filtered) ensures that any residual impact on model performance remains negligible, overcoming the limitations exposed in \cite{cao2020fltrust}. As shown in \cite{fourier}, this robust aggregation has an insignificant negative impact while maximizing model accuracy. Hence, as shown in Algorithm \ref{alg:methodology}, after obtaining the set of alleged benign clients, we apply the FFT for aggregating the weights and send these weights to the clients to continue the training.

\begin{algorithm}
\caption{FLAegis description}\label{alg:methodology}
\begin{algorithmic}[1]
\REQUIRE{$K$ set of clients, $(W_k)_{k\in K}$ weights of the clients}
\ENSURE{Aggregated Weights}\hfill \break
\textbf{Identification phase}
\FOR{$k \in K$}
\STATE $\Tilde{W_k} = SAX(W_k)$
\ENDFOR

\FOR{$k \in K$}
    \FOR{$l \in K$}
        \STATE $m_{kl} = cosine\_similarity(\Tilde{W_k},\Tilde{W_l})$
    \ENDFOR
\ENDFOR

\STATE $M=(m_{kl})_{k,l=1}^{|K|}$ similarity matrix

\STATE$(S_i)_1^L = Spectral\_Clustering(M)$

\IF{$L> 1$}
\STATE $S_1,S_2 = \text{K-Means}(M)$
\IF{$S_1 \geq S_2$}
    \STATE $B = S_1$ are the benign clients
\ELSE
    \STATE $B=S_2$ are the benign clients
\ENDIF
\ELSE 
\STATE $B=K$ are the benign clients
\ENDIF

\vspace{5pt}
\textbf{Mitigation phase}
\STATE $W = FFT((W_b)_{b\in B})$

\STATE Server sends $W$ to clients of $K$

\end{algorithmic}
\end{algorithm}

This two-stage defense approach forms the core of our proposed FLAegis framework. In the next section, we empirically evaluate its performance under a variety of poisoning attacks and compare it against state-of-the-art defenses.

\section{Experiments}\label{sec:Results}
In this section, we present a comprehensive evaluation of FLAegis under multiple poisoning attack scenarios. We first describe the experimental setup, including the model, dataset, and attack configurations. Then, we assess the performance of our approach in terms of both detection accuracy and final model robustness.

\subsection{Settings and Dataset}
For the evaluation of our two-phase defense strategy, we consider several federated scenarios running in a virtual machine with an Intel(R) Xeon(R) Silver 4214R CPU @ 2.40GHz with 32 cores and 96 GB of RAM. We use Flower \cite{flower} to implement the FL setup. All the parameters related to our federated scenario are summarized in Table \ref{tab:summary_fl}.

\begin{table}[]
\centering
\begin{tabular}{ll}
\hline
\textbf{Framework} & Flower            \\ \hline
\textbf{Model}     & CNN  \\ \hline
%\textbf{Aggregation function}     & FFT \\ \hline
\textbf{Clients}   & 50     \\ \hline
\textbf{Rounds}    & 50                \\ \hline
\textbf{Epochs}    & 1               \\ \hline
\end{tabular}
\caption{Summary of parameters for our federated scenario}
\label{tab:summary_fl}
\end{table}

The dataset used in this work is FEMNIST, a federated version of EMNIST \cite{cohen2017emnist} created by LEAF \cite{caldas2018leaf} available in their GitHub repository\footnote{\url{https://github.com/TalwalkarLab/leaf}}. FEMNIST is a dataset of handwritten characters used for image classification. LEAF takes the EMNIST dataset and divides it into 3550 non-iid clients. It contains 62 classes (the numbers from 0 to 9, and 52 letters in both upper and lower case). 

The model used for this dataset is a Convolutional Neural Network (CNN) composed of four hidden layers: three 8x8 convolutional layers with 8, 16, and 24 channels respectively, and a fully connected layer of 128 neurons. All hidden layers use the ReLu activation function, and the output layer has 62 neurons with softmax activation function. We use Adam as the optimizer with a learning rate of 0.001. Finally, the batch size is 64. In our case, we select 50 random clients that remain fixed throughout the experiments. The clients have an average of 1315 samples, 47 classes, and an entropy of 0.93. The entropy measures the balance between the classes within each client, and is calculated using the Shannon entropy \cite{bonachela2008entropy}. The malicious clients are chosen randomly from the same 50-client pool, and are re-sampled in each round. Additionally, we assume a scenario with no dropouts or \textit{stragglers} \cite{park2021sageflow}, meaning the nodes remain active throughout the entire training process.

\subsection{Evaluated Poisoning attacks }\label{desc_ataques}
The attacks considered are representative and impactful examples of poisoning strategies. We include one representative data poisoning technique and several local model poisoning attacks, covering both simple and sophisticated scenarios. From the data poisoning category, we select the label flipping attack, a widely used form of dirty-label poisoning. Dirty-label attacks are particularly relevant in FL since they do not require control over the training process, only over the labels. From the model poisoning category, we focus on various untargeted attacks, as untargeted manipulations have been shown to be more difficult to detect than targeted ones \cite{kiourti2020trojdrl}. We detail the attacks used in this work in Table \ref{tab:my_label}, and describe them as follows:

\begin{enumerate}
    \item \textbf{Label Flipping: }Label flipping \cite{lyu2020threats} is a type of data poisoning attack in which the labels of a client are systematically modified from a source class to another class, while leaving the training samples untouched.
    
    \item \textbf{LIE: }LIE (Little Is Enough) \cite{baruch2019little} is a model poisoning attack where small amounts of noise are added to each dimension of the average of their updates. Specifically, the malicious clients calculate the mean ($\nabla^M$) and standard deviation ($\sigma^M$) of their weights after the training round. Then, the malicious clients compute a coefficient $z$ based on the number of benign and malicious clients. Finally, the poisoned model update is given by $\nabla^M$ plus $z\sigma^M$, designed to remain below detection thresholds.
    
    \item \textbf{STATOPT: }STATOPT (Static Optimization) \cite{fang2020local} starts by computing the average of malicious clients' weights ($\nabla^M$) and derives a static malicious direction $\omega = -sign(\nabla^M)$. The poisoned update sent to the server is $-\gamma\omega$, where $\gamma$ is a variable chosen by the attackers. 

    \item \textbf{Mimic: } Mimic attack \cite{karimireddy2020byzantine} targets a benign client whose update has high variance, and all malicious clients replicate that update, thereby producing a bias in the global aggregation. Because the selected update originates from a benign client, the attack becomes harder to detect
    
    \item \textbf{Min-max: }The min-max attack was proposed in \cite{shejwalkar2021manipulating}. Its goal is to maximize the distance between malicious and benign updates while avoiding detection by the aggregation rule. In particular, malicious clients compute the average of their weights ($\nabla^M$). Then, they identify a perturbation direction ($\theta$) and scale it by a magnitude $\gamma$, producing the final malicious update $\nabla^M + \gamma\theta$. $\theta$ is an adversarially chosen direction in the model update space. The magnitude $\gamma$ is obtained by solving an optimization problem that ensures the resulting update remains within the statistical bounds of other malicious clients.
    
    \item \textbf{Min-sum: }  The min-sum attack is also proposed in \cite{shejwalkar2021manipulating}. It is similar to min-max as the same parameters are considered, the average $\nabla^M$, the direction $\theta$, and the distance $\gamma$. However, the optimization objective is to maximize the sum of distances between the perturbed update $\nabla^M + \gamma\theta$ and all other malicious updates, subject to a constraint on their total dispersion.
\end{enumerate}

\begin{table*}[ht!]
\centering
\begin{tabular}{ccccc}
\textbf{Attack} & \textbf{Type} & \textbf{Timing} & \textbf{Collude} & \textbf{Strategy} \\ \hline
\textbf{Label Flipping} & Data & Before training & No & Swap labels \\ \hline
\textbf{LIE} & Model & During training & Yes & Add noise scaled by standard deviation \\ \hline
\textbf{STATOPT} & Model & During training & Yes & Opposite-direction perturbation \\  \hline
\textbf{Mimic} & Model & During training & Yes & Replicate high-variance benign update \\ \hline
\textbf{Min-max} & Model & During training & Yes & Maximize distance under aggregation constraints \\ \hline
\textbf{Min-sum} & Model & During training & Yes & Optimize the distance of malicious weights based on the sum of the distances\\ \hline
\end{tabular}
\caption{Summary of the poisoning attacks considered in this work}
\label{tab:my_label}
\end{table*}

\subsection{Results}
 This subsection presents a comparative evaluation of several defense methods against the poisoning attacks introduced in Section \ref{desc_ataques}. Table \ref{tab:methods} summarizes the evaluated methods, which include our proposed approach (FLAegis), as well as baselines such as SignGuard, FedDMC, and LoMar, described in Section \ref{sec:related_methods_for_comparison}. First, we evaluate the detection rate of these methods, focusing on their effectiveness in identifying malicious clients. Next, we analyze the performance of the different methods in terms of accuracy.  Additionally, we compare their performance when combining them with the FFT-based aggregation function proposed in our previous work \cite{fourier} (i.e., replacing their default FedAvg aggregation). Lastly, to check the validity of the statement made in \cite{xu2022byzantine, wan2022shielding}, we carry out an ablation study, in which we analyze the impact of SAX SAX preprocessing and FFT aggregation on the performance of FLAegis.

\begin{table*}[]
\centering
\begin{tabular}{|>{\centering\arraybackslash}p{1.4cm} 
                |>{\centering\arraybackslash}p{1.4cm} 
                |>{\centering\arraybackslash}p{2.2cm} 
                |>{\centering\arraybackslash}p{1.4cm} 
                |>{\centering\arraybackslash}p{3.8cm} 
                |>{\centering\arraybackslash}p{1.5cm} 
                |>{\centering\arraybackslash}p{1.5cm}|}
\hline
\textbf{Method} & \textbf{Input type} & \textbf{Preprocessing} & \textbf{Similarity matrix} & \textbf{Client classification} & \textbf{Historical information} & \textbf{Robust aggregation} \\ \hline
\textbf{SignGuard} & Gradients & No & No & Intersection of two filters & No & No \\ \hline
\textbf{FedDMC} & Weights & PCA & No & Clustering via binary tree-based clustering with noise (BTBCN) & Yes & No \\ \hline
\textbf{LoMar} & Weights & k-NN + estimated distribution & No & Based on threshold & No & No \\ \hline
\textbf{FLAegis} & Weights & SAX & Yes & Spectral clustering & No & Yes \\ \hline
\end{tabular}

\caption{Comparison of the evaluated defense methods}
\label{tab:methods}
\end{table*}

\subsubsection{\textbf{Malicious client detection accuracy}}\label{sec:detecction_accuracy}
\begin{figure*} [!ht]
	\centering
		\includegraphics[width=2.2\columnwidth]{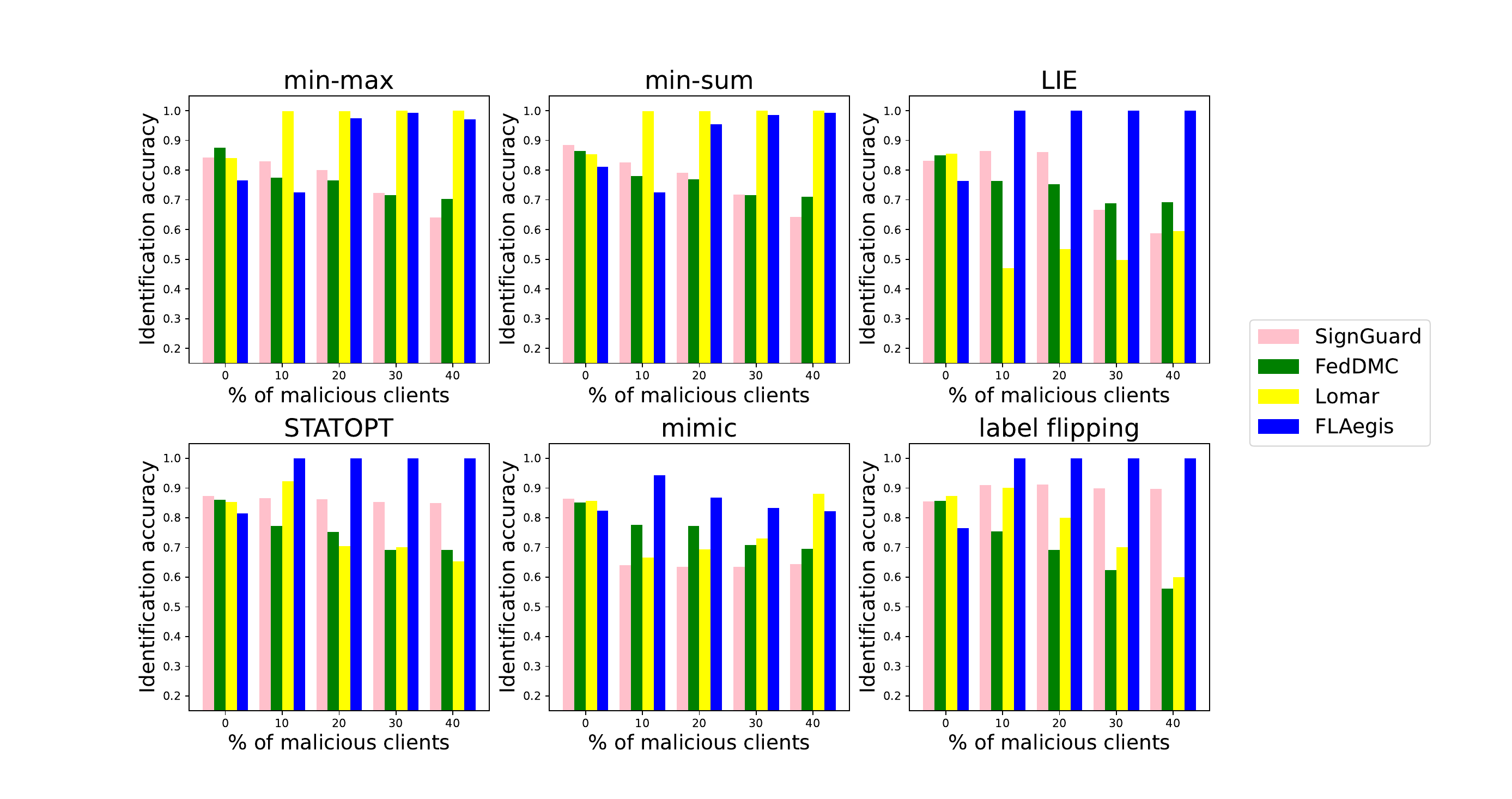}
	\caption{Detection accuracy of the different methods against the attacks}
	\label{fig:barras}
\end{figure*}

\hfill \newline
The main objective of FLAegis and the other defenses described in Section \ref{sec:Related} is to identify malicious clients in order to preserve the aggregated model's integrity. This identification involves accurately detecting adversarial participants while minimizing false positives among benign clients. Figure \ref{fig:barras} compares the detection performance of FLAegis, SignGuard, FedDMC, and LoMar measured as the proportion of correctly classified clients, including both benign and malicious ones. SignGuard maintains a stable performance under the STATOPT and label flipping attacks, with detection accuracy ranging from 0.85 to 0.9. However, its accuracy degrades notably under the other attacks. Specifically, for the LIE attack, SignGuard achieves 0.85 accuracy when there are between 0 and 20\% of malicious clients; beyond that, the accuracy drops to 0.6.  In min-max and min-sum, detection accuracy starts at 0.83 and 0.88 respectively, and drops to 0.64. For FedDMC, performance remains between 0.8 to 0.7 across most attacks, though it drops to 0.6 for label flipping. Regarding LoMar, it performs best under min-max and min-sum. However, its effectiveness drops considerably for the remaining attacks. In particular, it shows a decreasing trend under label flipping, and achieves poor results under LIE, with detection rates between 0.4 and 0.6. Finally, FLAegis consistently outperforms the other methods. In the absence of malicious clients (0\%), detection accuracy remains close to 0.8 across all scenarios. When adversaries are present, detection reaches 100\% for LIE, STATOPT, and label flipping. Under the mimic attack, accuracy declines from 0.94 to 0.8. This occurs because malicious clients imitate a benign client with high weight variance; after applying SAX, such outliers occasionally bypass clustering detection. For the most sophisticated min-sum and min-max attacks, detection remains near perfect when the proportion of malicious clients exceeds 20\%, but drops to approximately 0.79 at 10\%. This is due to a lower density of adversaries and the subtle placement of poisoned updates close to benign ones in the parameter space, which hinders spectral clustering. This motivated the integration of FFT as a final defense layer, ensuring robustness even when detection is imperfect. As discussed in the following section, this addition mitigates occasional detection errors and ensures consistent model accuracy across all attack scenarios.

\subsubsection{\textbf{Model accuracy under adversarial conditions}}
\begin{figure*} [!ht]
	\centering
		\includegraphics[width=2.2\columnwidth]{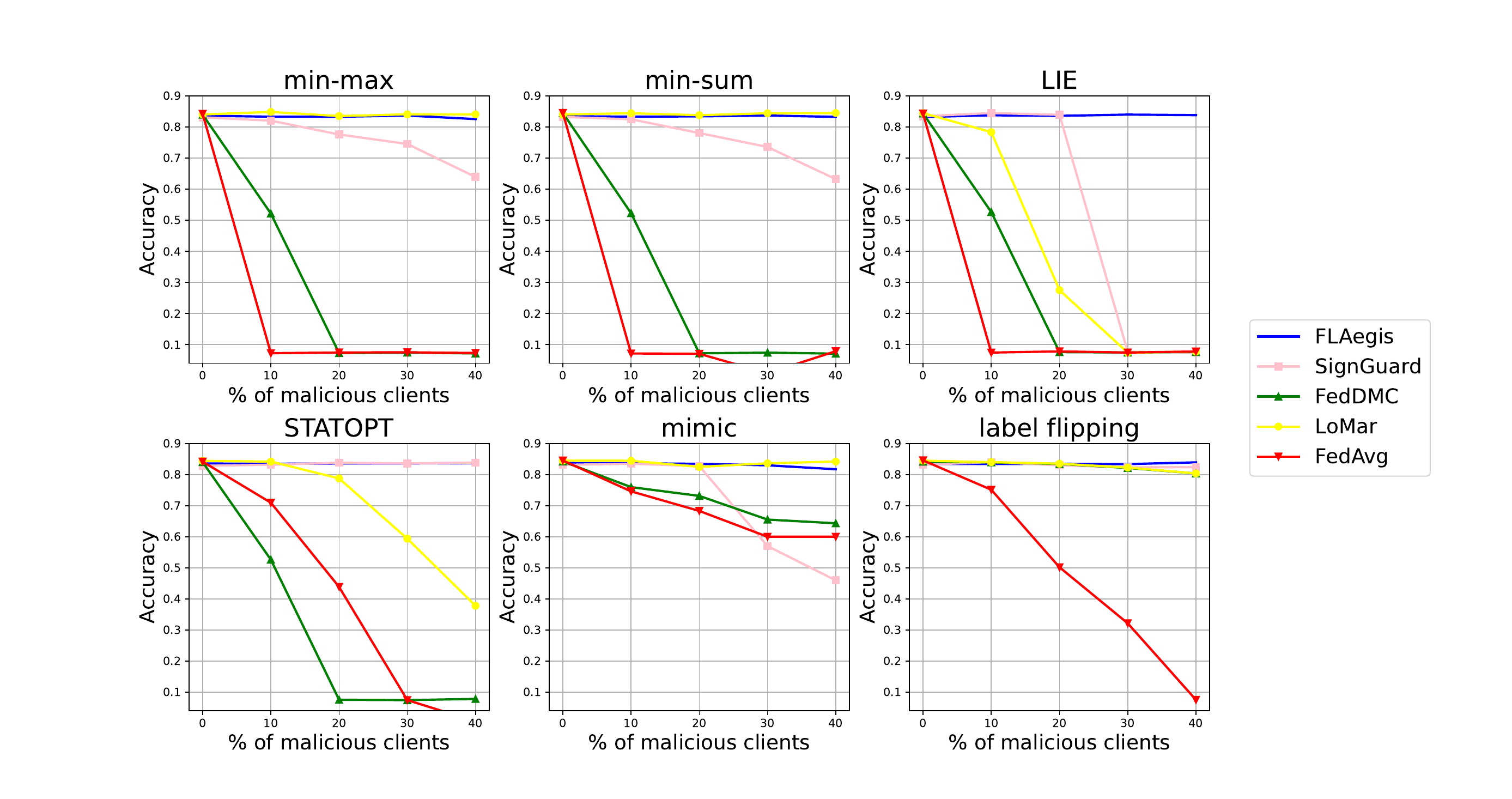}
	\caption{Comparison of our method with FedAvg, SignGuard, FedDMC and LoMar}
	\label{fig:graficas}
\end{figure*}

\hfill \newline
In this section, we evaluate the final accuracy obtained by the different methods against the attacks to assess their ability to maintain performance in adversarial scenarios using the FEMNIST dataset. First, in Fig. \ref{fig:graficas}, we compare the performance of our method to the original versions of SignGuard, FedDMC, and LoMar, as well as FedAvg without any defense system. In this figure, the accuracy of FedAvg without malicious clients (0.845) serves as a reference baseline, representing the ideal performance in a clean setting. Under adversarial conditions, however, FedAvg generally provides the worst results, except in STATOPT, where the poor detection rate of FedDMC causes it to perform even worse. Looking at the performance of the other defense methods, the results are consistent with those obtained in Fig. \ref{fig:barras}. In particular, SignGuard is capable of resisting STATOPT, label flipping, and LIE and mimic up to a malicious client proportion of 20\%, but for min-max and min-sum, it has a rapid decline. Regarding FedDMC, except for label flipping, its accuracy drops quickly as the proportion of malicious clients increases across the other attacks. LoMar performs best among these three methods, except in LIE and STATOPT attacks. In LIE and STATOPT, the detection accuracy was lower than in the others, which explains the decrease in accuracy. In the rest of the attacks, LoMar achieves accuracies between 0.83 and 0.84. Finally, our method consistently reaches the best performance overall, with accuracies around 0.83, except for the mimic attack. Although detection is less precise in this case, the vast majority of malicious clients are still filtered, resulting in an accuracy of 0.81 even when 40\% of clients are adversarial.

We now analyse the limitations observed in the remaining defense methods. In SignGuard, its original evaluation was conducted under an IID setting, which partially explains the low performance obtained against more complex attacks. For FedDMC, the PCA-based feature reduction is constrained by the model architecture. In our setup, reducing to three features significantly compromised the method’s discriminative capability. The effectiveness of LoMar, although achieving solid results, depends heavily on the accuracy of the threshold derived from Equation~\ref{eq:lomar2}, which proved less reliable under LIE and STATOPT attacks. In general, the results of Fig. \ref{fig:graficas} and Fig. \ref{fig:barras} are correlated: a strong detection mechanism leads to a notable final accuracy.

\begin{figure*} [!ht]
	\centering
		\includegraphics[width=2.2\columnwidth]{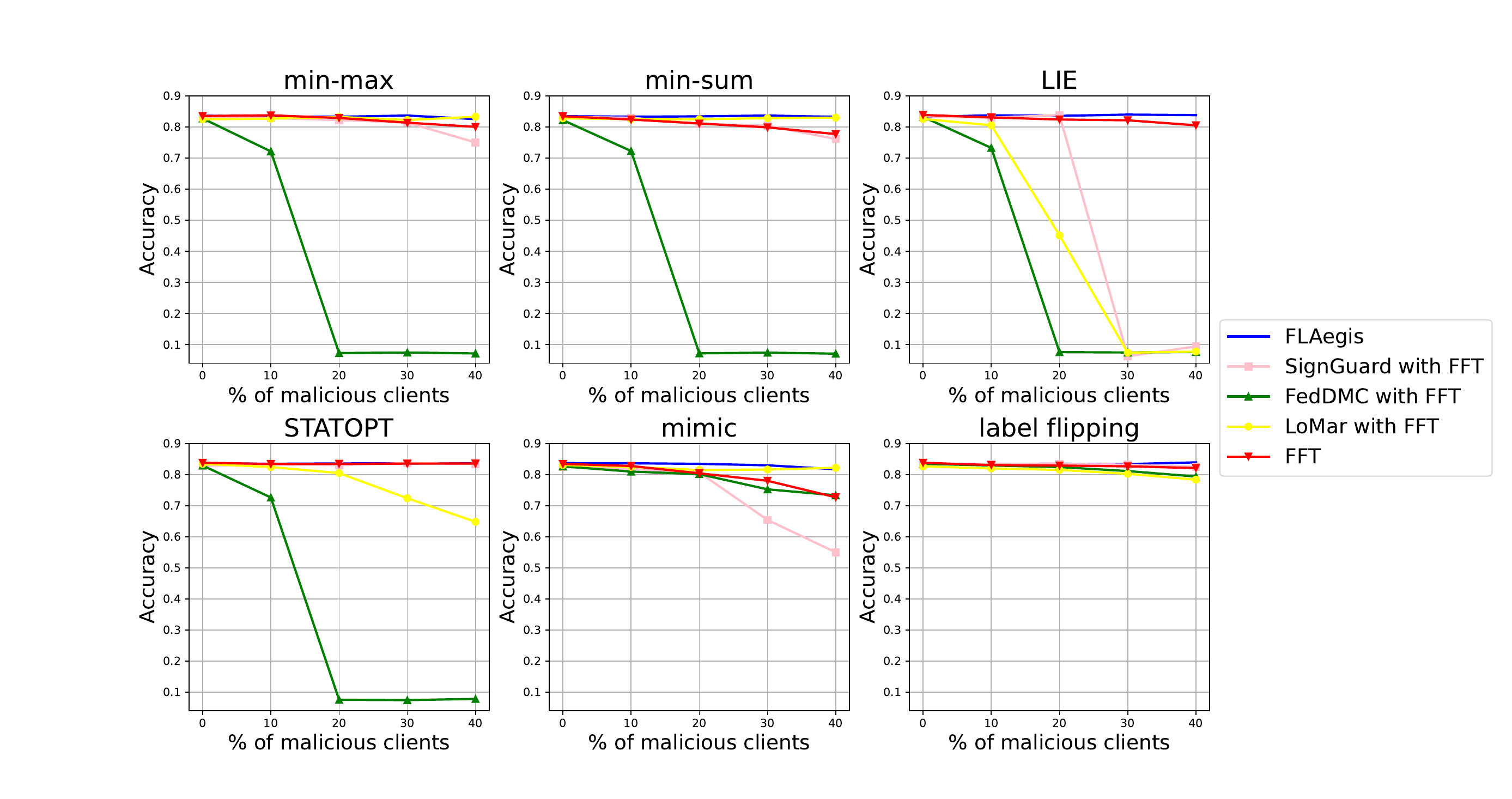}
	\caption{Comparison of our method, the FFT, and the FFT-extended versions of SignGuard, FedDMC and LoMar}
	\label{fig:graficasfft}
\end{figure*}

To further mitigate the influence of undetected adversaries, FLAegis applies FFT as a robust aggregation function after the identification phase. For comparison under equal conditions, we extend SignGuard, FedDMC and LoMar by incorporating FFT as their aggregation mechanism. Additionally, we evaluate FFT in isolation to demonstrate that the identification phase employed by FLAegis is a necessary component, rather than a redundant step in the presence of a robust aggregator. Figure \ref{fig:graficasfft} presents the performance of all methods under the various attack scenarios. Although the differences are lower compared to Figure \ref{fig:graficas}, our method continues to achieve superior performance. FFT is the second best method, but in min-max, min-sum, and LIE attacks, as the proportion of malicious clients increases, its accuracy drops to 0.78, as expected based on the findings in \cite{fourier}. Hence, the implementation of the identification phase is justified as FFT alone is insufficient to fully defend against sophisticated attacks. When FFT is combined with SignGuard, FedDMC, and LoMar, the results show improved performance (Figure \ref{fig:graficas}), as the impact of undetected malicious clients is reduced by the robust aggregation. However, despite this improvement, a decreasing trend in accuracy persists as the proportion of malicious clients increases, highlighting the superior resilience of FLAegis under adversarial conditions. Furthermore, as shown in Figure \ref{fig:barras}, as the malicious presence increases, the detection accuracy of SignGuard, FedDMC, and LoMar decreases, which leads to misclassification of benign clients as malicious, and malicious as benign, increasing the benign-malicious ratio during aggregation and amplifying the performance gap between our method and the others. Hence, having a high detection accuracy is essential for achieving a robust final model accuracy.

\subsubsection{\textbf{Impact of SAX and FFT on FLAegis}}\hfill \newline
This section evaluates the importance of the two main components of our method, SAX and FFT, by analyzing their individual impact when they are removed. First, motivated by \cite{xu2022byzantine, wan2022shielding}, which advocate for data preprocessing before applying the cosine similarity, we show the results of our method when SAX is omitted and cosine similarity is applied directly. Next, we evaluate the performance of FLAegis without its mitigation phase by replacing FFT with FedAvg, to assess the benefit of using FFT and the synergy between identification and aggregation. In Table \ref{tab:det_SAX}, we show the detection accuracy of the identification phase of FLAegis with and without SAX. As shown,  for 0\% malicious clients and under LIE, STATOPT, and label flipping attacks, the results remain similar. However, in min-max, min-sum, and mimic, SAX introduces noticeable differences. In particular, for min-max and min-sum, detection accuracy improves significantly when SAX is used, especially as malicious presence increases. In the case of the mimic attack, since the attack replicates high-variance benign weights, SAX may assign similar symbolic representations, reducing clustering effectiveness and causing some misclassifications. This reflects a relative weakness of SAX when dealing with adversaries that mimic benign clients with high weight variance. However, as shown in the accuracy evaluation, the integration of the FFT in the aggregation phase compensates for these cases, mitigating their impact on the global model performance.

Next, in Table \ref{tab:comp_SAX}, we compare the impact of this detection accuracy on the final accuracy. As shown, the final accuracy of our method without SAX decreases as the malicious presence increases in min-max and min-sum attacks, while in the mimic attack it improves slightly. Regarding the importance of implementing the FFT, Table \ref{tab:comp_SAX} also compares FLAegis without FFT, i.e., using FedAvg instead. At 0\%, both variants behave similarly. Nonetheless, at 10\%, thanks to the FFT, the original FLAegis achieves better results in scenarios where detection rate is lower than in the other percentages, as we explained in Section \ref{sec:detecction_accuracy}. Additionally, as in the mimic attack, imperfect detection leads to lower accuracy without FFT as malicious clients increase. However, for the other attacks, in which the identification rate is perfect,  FedAvg slightly outperforms FFT, consistent with the results provided in our previous work \cite{fourier}.

Based on the analysis above, we claim that while omitting components may occasionally yield marginal gains, the full FLAegis pipeline consistently outperforms partial versions, confirming the necessity of both SAX and FFT across all attack scenarios.

\begin{table*}[]
\begin{tabular}{l|cc|cc|cc|cc|cc|}
\cline{2-11}
                                              & \multicolumn{2}{c|}{\textbf{0 \%}}                                                                       & \multicolumn{2}{c|}{\textbf{10 \%}}                                                                      & \multicolumn{2}{c|}{\textbf{20 \%}}                                                                      & \multicolumn{2}{c|}{\textbf{30 \%}}                                                                      & \multicolumn{2}{c|}{\textbf{40 \%}}                                                                      \\ \cline{2-11} 
\multicolumn{1}{c|}{}                         & \multicolumn{1}{c|}{\textbf{FLAegis}} & \textbf{\begin{tabular}[c]{@{}c@{}}FLAegis\\  w/o SAX\end{tabular}} & \multicolumn{1}{c|}{\textbf{FLAegis}} & \textbf{\begin{tabular}[c]{@{}c@{}}FLAegis\\  w/o SAX\end{tabular}} & \multicolumn{1}{c|}{\textbf{FLAegis}} & \textbf{\begin{tabular}[c]{@{}c@{}}FLAegis\\  w/o SAX\end{tabular}} & \multicolumn{1}{c|}{\textbf{FLAegis}} & \textbf{\begin{tabular}[c]{@{}c@{}}FLAegis \\ w/o SAX\end{tabular}} & \multicolumn{1}{c|}{\textbf{FLAegis}} & \textbf{\begin{tabular}[c]{@{}c@{}}FLAegis \\ w/o SAX\end{tabular}} \\ \hline
\multicolumn{1}{|l|}{\textbf{Min-max}}         & \multicolumn{1}{c|}{0.7663}           & 0.7366                                                              & \multicolumn{1}{c|}{0.7258}           & 0.7258                                                              & \multicolumn{1}{c|}{0.9743}           & 0.9117                                                              & \multicolumn{1}{c|}{0.9921}           & 0.9444                                                              & \multicolumn{1}{c|}{0.9716}           & 0.9286                                                              \\ \hline
\multicolumn{1}{|l|}{\textbf{Min-sum}}         & \multicolumn{1}{c|}{0.8105}           & 0.8057                                                              & \multicolumn{1}{c|}{0.7254}           & 0.7345                                                              & \multicolumn{1}{c|}{0.9536}           & 0.9241                                                              & \multicolumn{1}{c|}{0.9864}           & 0.9335                                                              & \multicolumn{1}{c|}{0.9936}           & 0.9271                                                              \\ \hline
\multicolumn{1}{|l|}{\textbf{LIE}}            & \multicolumn{1}{c|}{0.7634}           & 0.8169                                                              & \multicolumn{1}{c|}{1}                & 1                                                                   & \multicolumn{1}{c|}{1}                & 1                                                                   & \multicolumn{1}{c|}{1}                & 1                                                                   & \multicolumn{1}{c|}{1}                & 1                                                                   \\ \hline
\multicolumn{1}{|l|}{\textbf{STATOPT}}        & \multicolumn{1}{c|}{0.8155}           & 0.7737                                                              & \multicolumn{1}{c|}{1}                & 1                                                                   & \multicolumn{1}{c|}{1}                & 1                                                                   & \multicolumn{1}{c|}{1}                & 1                                                                   & \multicolumn{1}{c|}{1}                & 1                                                                   \\ \hline
\multicolumn{1}{|l|}{\textbf{Mimic}} & \multicolumn{1}{c|}{0.812}            & 0.8025                                                              & \multicolumn{1}{c|}{0.9416}                & 1                                                                   & \multicolumn{1}{c|}{0.8676}                & 1                                                                   & \multicolumn{1}{c|}{0.8324}                & 1                                                                   & \multicolumn{1}{c|}{0.8216}                & 1                                                                   \\ \hline
\multicolumn{1}{|l|}{\textbf{Label Flipping}} & \multicolumn{1}{c|}{0.765}            & 0.7815                                                              & \multicolumn{1}{c|}{1}                & 1                                                                   & \multicolumn{1}{c|}{1}                & 1                                                                   & \multicolumn{1}{c|}{1}                & 1                                                                   & \multicolumn{1}{c|}{1}                & 1                                                                   \\ \hline
\end{tabular}
\caption{Detection accuracy of FLAegis with and without SAX under different attack types and varying percentages of malicious clients}
\label{tab:det_SAX}
\end{table*}

\begin{table*}[]
\centering
\centering
\begin{tblr}{
  colspec = {cc|c|c|c|c|c|c},
  row{2-16} = {m},
  cell{2,5,8,11,14}{1} = {r=3}{m},      % el porcentaje ocupa 3 filas
  vline{1,2}={2-22}{},
  vline{9}={1-22}{}, %3,4,6,7,9,10,,12,13,15,16,18,19,21
  hline{3,4,6,7,9,10,12,13,15,16,18,19,21} = {2-8}{},
  hline{1,2} = {3-8}{},      % línea bajo el encabezado (entre fila 1 y 2)
  hline{5,8,11,14,17} = {-}{}
}
  & & \textbf{minmax} & \textbf{minsum} & \textbf{LIE} & \textbf{STATOPT} & \textbf{Mimic} & \textbf{Label Flipping} \\
\hline
\textbf{0 \%}  & \textbf{FLAegis}                         & 0.8361 & 0.8347 & 0.8319 & 0.8356 & 0.8340 & 0.8341 \\
               & \makecell{\textbf{FLAEgis}\\\textbf{w/o SAX}} & 0.8352 & 0.8341 & 0.8321 & 0.8314 & 0.8310 & 0.8327 \\
               & \makecell{\textbf{FLAEgis}\\\textbf{w/o FFT}} & 0.8371 & 0.8393 & 0.8391 & 0.8423 & 0.8361 & 0.8437 \\
\hline
\textbf{10 \%} & \textbf{FLAegis}                         & 0.8330 & 0.8329 & 0.8375 & 0.8346 & 0.8360 & 0.8340 \\
               & \makecell{\textbf{FLAEgis}\\\textbf{w/o SAX}} & 0.8312 & 0.8294 & 0.8317 & 0.8310 & 0.8411 & 0.8339 \\
               & \makecell{\textbf{FLAEgis}\\\textbf{w/o FFT}} & 0.8086 & 0.8046 & 0.8392 & 0.8380 & 0.7753 & 0.8381 \\
\hline
\textbf{20 \%} & \textbf{FLAegis}                         & 0.8328 & 0.8341 & 0.8357 & 0.8357 & 0.8345 & 0.8346 \\
               & \makecell{\textbf{FLAEgis}\\\textbf{w/o SAX}} & 0.8246 & 0.8224 & 0.8337 & 0.8303 & 0.8410 & 0.8305 \\
               & \makecell{\textbf{FLAEgis}\\\textbf{w/o FFT}} & 0.8366 & 0.8395 & 0.8410 & 0.8376 & 0.7256 & 0.8422 \\
\hline
\textbf{30 \%} & \textbf{FLAegis}                         & 0.8366 & 0.8366 & 0.8397 & 0.8359 & 0.8297 & 0.8336 \\
               & \makecell{\textbf{FLAEgis}\\\textbf{w/o SAX}} & 0.8103 & 0.8058 & 0.8341 & 0.8305 & 0.8397 & 0.8302 \\
               & \makecell{\textbf{FLAEgis}\\\textbf{w/o FFT}} & 0.8416 & 0.8402 & 0.8361 & 0.8369 & 0.6540 & 0.8383 \\
\hline
\textbf{40 \%} & \textbf{FLAegis}                         & 0.8253 & 0.8325 & 0.8380 & 0.8366 & 0.8152 & 0.8392 \\
               & \makecell{\textbf{FLAEgis}\\\textbf{w/o SAX}} & 0.7752 & 0.7841 & 0.8304 & 0.8326 & 0.8327 & 0.8310 \\
               & \makecell{\textbf{FLAEgis}\\\textbf{w/o FFT}} & 0.8388 & 0.8414 & 0.8428 & 0.8484 & 0.5924 & 0.8491 \\
\end{tblr}
\caption{Impact of disabling SAX or FFT on the final accuracy of FLAegis across different attack types and adversarial ratios}
\label{tab:comp_SAX}
\end{table*}

\subsubsection{\textbf{Overall performance analysis}}\hfill \newline
Based on the previous analysis and the results of Figures \ref{fig:barras}, \ref{fig:graficas}, and \ref{fig:graficasfft}, it becomes evident that correctly identifying the malicious weights is critically important. Even when a robust aggregation function is applied, the results can vary significantly, particularly under a high percentage of malicious clients. In this context, FLAegis consistently outperforms existing state-of-the-art defenses due to its combination of strong identification capability and the use of FFT in the aggregation phase, which acts as a safeguard when the detection is not entirely accurate. Table \ref{tab:comp_SAX} further demonstrates that each component of the method contributes meaningfully to its robustness. Hence, FLAegis is designed to operate effectively in more challenging and non-IID settings, unlike previous works. In our experimental setup, the performance of other techniques degraded significantly for certain attacks, even when enhanced with FFT, underscoring the need for integrated, adaptive defense mechanisms such as those employed by FLAegis.

\section{Conclusions and Future Work}\label{sec:Conclusions}
In this study, we have introduced FLAegis, a novel approach aimed at safeguarding federated models by detecting and excluding malicious clients from the aggregation process. Our methodology revolves around identifying these malicious clients through the construction of a similarity matrix based on their SAX-transformed representations. Additionally, as a final security layer, we incorporate an FFT-based aggregation function to counter the influence of clients that evade initial detection. This dual-layer design enhances resilience under high adversarial presence, where detection may not always be perfect and robust aggregation alone tends to degrade. Comparatively, our method 
outperforms state-of-the-art approaches, particularly under sophisticated untargeted attacks like min-max and min-sum, where other methods struggle. As part of future work, we aim to improve the identification phase to eliminate reliance on FFT entirely, further simplifying the pipeline and improving accuracy in scenarios with minimal or well-separated adversarial influence.
%\input{taxonomy}
%\input{challenges}
%\input{conclusions}

\section*{Acknowledgment}
This study was supported by grant RYC2023-043553-I, funded by MICIU/AEI/10.13039/501100011033 and ESF+. It has also been funded by CNS2023-145059 (INTEGRATOR) funded by MICIU/AEI/10.13039/501100011033 and the European Union NextGenerationEU/PRTR, by the project PCI2023-145989-2 (REMINDER) funded by MICIU/AEI/10.13039/501100011033 and by the European Union NextGenerationEU/PRTR, by the project PID2023-148104OB-C43 (ONOFRE4) funded by  MICIU/AEI/10.13039/501100011033 and FEDER/UE, by the Leonardo Grant 2023 to Researchers and Cultural Creators from the BBVA Foundation, and by the European Commission through the HORIZON-MSCA-2021-SE-01-01 project Cloudstars (g.a. 101086248).

\bibliographystyle{IEEEtran}
\bibliography{biblio}
\appendices

%\end{IEEEbiography}

%\begin{IEEEbiographynophoto}{José L. Hernández-Ramos} received the  Ph.D. degree in computer science from the University of Murcia (UMU), Spain. He is currently a Marie Sklodowska-Curie Postdoc Fellow at UMU working on the topic of federated learning and intrusion detection. Before that, he was a Scientific Project Officer with the Joint Research Centre, European Commission. He has participated in different European research projects, such as SocIoTal, SMARTIE, and SerIoT. He has published more than 60 peer-reviewed papers. His research interests include application of security and privacy mechanisms in the Internet of Things and transport systems scenarios, including blockchain and machine learning.\end{IEEEbiographynophoto}

\begin{IEEEbiography}[{\includegraphics[width=1in,height=1.25in,clip,keepaspectratio]{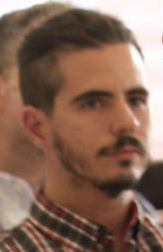}}]{Enrique Mármol Campos}
 is a postdoctoral researcher at the university of Murcia. He graduated in Mathematics in 2018. Then, in 2019, he finished the M.S. in advanced math, in the specialty of operative research and statistic, at the university of Murcia. And in 2024, he got a Ph.D. in Computer Science. He is currently researching on federated learning applied to cybersecurity in IoT devices.
\end{IEEEbiography}

\begin{IEEEbiography}[{\includegraphics[width=1in,height=1.25in,clip,keepaspectratio]{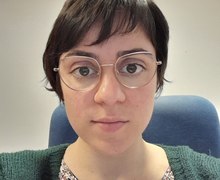}}]{Aurora Gonzalez Vidal}
 graduated in Mathematics from the University of Murcia in 2014. In 2015 she got a fellowship to work in the Statistical Division of the Research Support Service, where she specialized in Statistics and Data Analysis. Afterward, she studied a Big Data Master. In 2019, she got a Ph.D. in Computer Science. Currently, she is a Ramon y Cajal postdoctoral researcher at the University of Murcia. She has collaborated in several national and European projects such as ENTROPY, IoTCrawler, and DEMETER. Her research covers machine learning in IoT-based environments, missing values imputation, and time-series segmentation. She is the president of the R Users Association UMUR.
\end{IEEEbiography}

\begin{IEEEbiography}[{\includegraphics[width=1in,height=1.25in,clip,keepaspectratio]{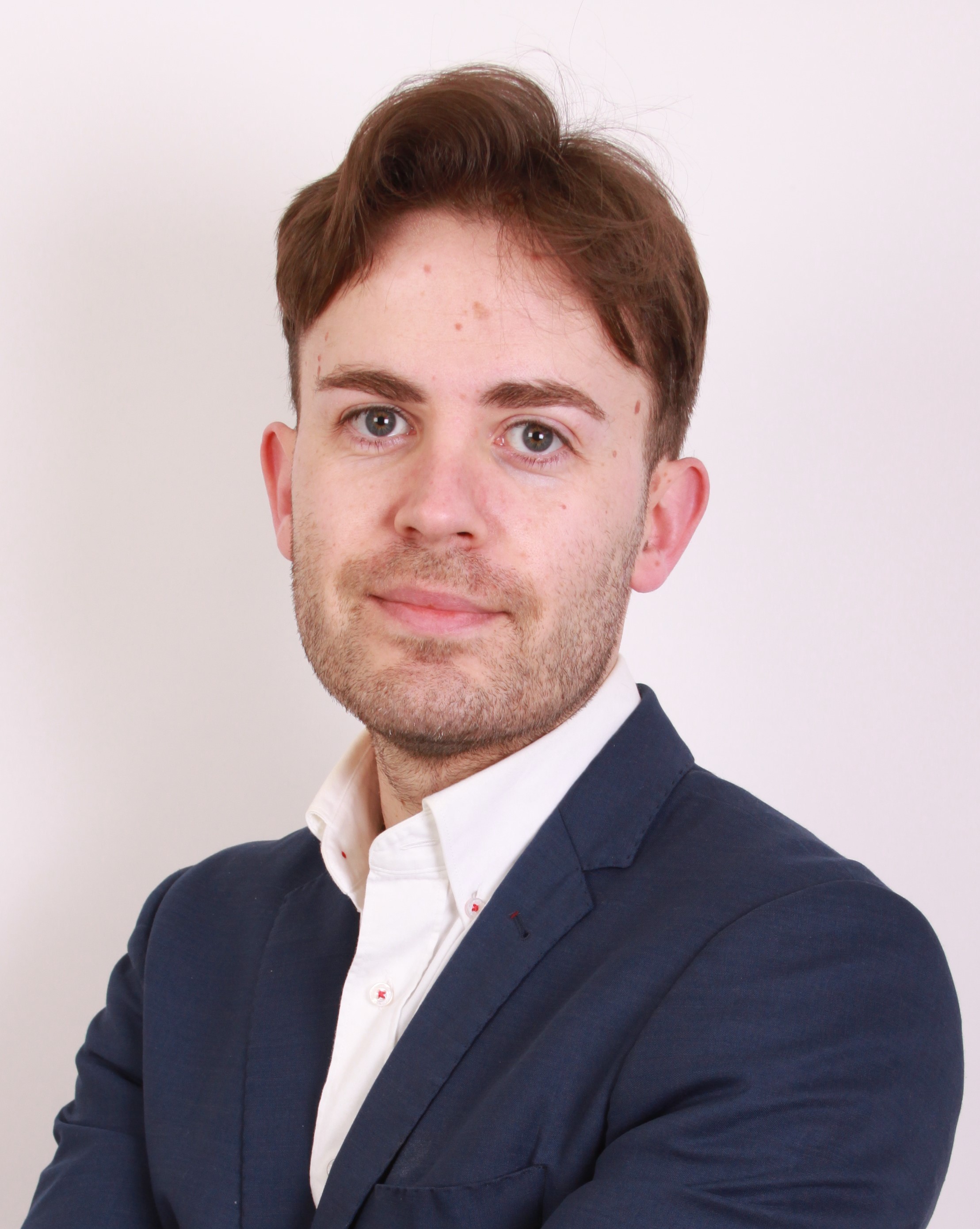}}]{José Luis Hernández Ramos}
 received the Ph.D. degree in computer science from the University of Murcia, Spain. He is an associate professor at the same university, and previously he was a Marie Sklodowska-Curie Postdoctoral Fellow. He worked as a Scientific Project Officer at the European Commission. He has participated in different European research projects, such as SocIoTal, SMARTIE, and SerIoT, and published more than 60 peer-reviewed papers. His research interests include the application of AI techniques in cybersecurity and data protection. He has served as a technical program committee and chair member for several international conferences, and editor in different journals.
\end{IEEEbiography}

\begin{IEEEbiography}[{\includegraphics[width=1in,height=1.25in,clip,keepaspectratio]{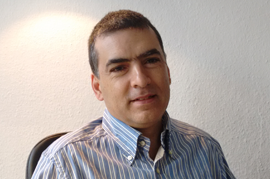}}]{Antonio Skarmeta}
 is a full professor at the University of Murcia, the Department of Information and Communications Engineering. His research interests are the integration of security services, identity, the Internet of Things, and smart cities. Skarmeta received a Ph.D. in computer science from the University of Murcia. He has published more than 200 international papers and been a member of several program committees.
\end{IEEEbiography}

\end{document}